\documentclass{article}

     \usepackage[preprint]{neurips_2023}

\usepackage[utf8]{inputenc} 
\usepackage[T1]{fontenc}    
\usepackage{hyperref}       
\usepackage{url}            
\usepackage{booktabs}       
\usepackage{amsfonts}       
\usepackage{nicefrac}       
\usepackage{microtype}      
\usepackage{xcolor}         
\usepackage{graphicx}
\usepackage[]{algorithmic} 
\usepackage[ruled,linesnumbered]{algorithm2e}

\title{Talking Models: Distill Pre-trained Knowledge to Downstream Models via Interactive Communication}

\author{%
    Zhe Zhao$^1$, Qingyun Liu$^1$, Huan Gui$^1$, Bang An$^2$\thanks{This work was done when Bang was an intern at Google.},\ Lichan Hong$^1$, Ed H. Chi$^1$ \\
      $^1$Google DeepMind,\  \texttt{\{zhezhao,qyl,hgui,lichan,edchi\}@google.com}\\
      $^2$University of Maryland,\ \texttt{bangan@umd.edu} \\
}

\begin{document}

\maketitle


\begin{abstract}
Many recent breakthroughs in machine learning have been enabled by the pre-trained foundation models. By scaling up model parameters, training data, and computation resources, foundation models have significantly advanced the state-of-the-art in many applications.  However, it is still an open question of how to use these models to perform downstream tasks efficiently. 
Knowledge distillation (KD) has been explored to tackle this challenge. KD is a technique that transfers knowledge from a large teacher model to a smaller student model. While KD has been successful in improving student model performance, recent research has discovered that a powerful teacher does not necessarily lead to a powerful student, due to their huge capacity gap. In addition, the potential distribution shifts between the pre-training data and downstream tasks can make knowledge transfer in KD sub-optimal for improving downstream task performance.

In this paper, we extend the knowledge distillation paradigm by introducing an interactive communication process to help student models of downstream tasks learn effectively from pre-trained foundation models. Our design is inspired by the way humans learn from teachers who can explain knowledge in a way that meets the students' needs. Specifically, we let each model (i.e., student and teacher) train two components: (1) an encoder which encodes the model's hidden states to a message in a shared message space and (2) a decoder which decodes any message to its own hidden states. With encoder and decoder, not only can the teacher model transfer rich information by encoding its hidden states to messages, but also the student model can send messages with information of downstream tasks to teacher so that the teacher can interpret and generate responses. With this interactive communication process, knowledge passing from teacher to student can be tailored to the student's model capacity and downstream tasks' distributions. We conducted experiments on benchmark datasets for computer vision and recommendation tasks to show that our communication mechanism outperforms state-of-the-art distillation techniques. 
\end{abstract}

\section{Introduction}

Scaling up machine learning models has been shown to improve performance in many applications, including Natural Language Processing (NLP) \citep{chowdhery2022palm}, Computer Vision (CV) \citep{yuan2021florence} and Information Retrieval tasks \citep{tay2022transformer}. For example, the number of parameters of newly developed language models for NLP tasks has grown from hundreds of millions to hundreds of billions in the past few years \citep{zhao2023survey}. By scaling both model size and training data, researchers have found that not only can large models reduce training loss \citep{kaplan2020scaling}, they can also show emergent abilities to solve tasks that smaller models cannot solve, such as in-context learning and instruction following \citep{wei2022emergent}. Similarly, in CV, large models can improve performance of multiple tasks of different datasets \citep{dosovitskiy2020image}. The training of these large models can be extremely costly, but once they are trained, they can be stored as pre-trained foundation models to be fine-tuned for downstream tasks \citep{raffel2020exploring, pretrainimagenet21k}. 

As the size of these foundation models grows bigger, the inference cost of the fine-tuned models also grows. On the contrary, for many real-world machine learning applications, researchers have built their own sophisticated but efficient models, such as real-time traffic sign detection \citep{bousarhane2021road}, or recommendation and retrieval \citep{zhao2019recommending}. Without techniques that can significantly bring down the inference cost of the fine-tuned models, it is challenging to apply foundation models to a wide range of applications with extensive inference requests and strict constrains on latency.

Existing efforts in reducing the inference cost of large machine learning models include model pruning \citep{frantar2022optimal}, quantization \citep{liang2021pruning} and knowledge distillation \citep{hinton2015distilling}. Model pruning \citep{frantar2022optimal} techniques prune learned weights based on their importance, and quantization \citep{dettmers2022llm} techniques approximate learned weights with data-types of low-bit-width. While both techniques provide a trade-off between efficiency and quality, they either need to be carefully designed for specific model architectures, or would degrade the quality of the model regardless of the downstream tasks. When transferring knowledge of pre-trained foundation models to perform downstream tasks, it is desirable to have flexibility for the downstream models adopting different model architectures, such as decoders from the pre-trained model. Therefore, knowledge distillation techniques, which transfer knowledge from one model to another by using model outputs (e.g., predictions), can provide such flexibility \citep{gou2021knowledge}. In this paper, we explore the direction of distilling the knowledge from pre-trained foundation models (teacher) to smaller models for downstream tasks (students) by knowledge distillation. 

Some very recent research \citep{hsieh2023distilling} successfully distilled large language model (LLM) to smaller models for downstream tasks, by using LLM generating rationale of the labels for the downstream models. As this research is specific to NLP applications and cannot be easily extended to other machine learning applications, these exist two additional challenges on applying KD to distill pre-trained knowledge to improve downstream models. Firstly, recent research discovers that a huge capacity gap between teacher and student will hurt KD \citep{wang2022efficient, huang2022knowledge}. Therefore it becomes difficult to apply KD to improve smaller downstream models by learning from much larger pre-trained models. Secondly, downstream tasks can have different distributions from the pre-trained data. Therefore knowledge from the pre-trained model may not be directly useful to downstream tasks. Although some recent study found that distillation from a fine-tuned teacher can improve the performance of the student model \citep{wu2023distilling}, fine-tuning pre-trained foundation models is very costly and cannot scale to many downstream tasks. 

\begin{figure}[t!]
    \centering
    \includegraphics[width=0.99\linewidth]{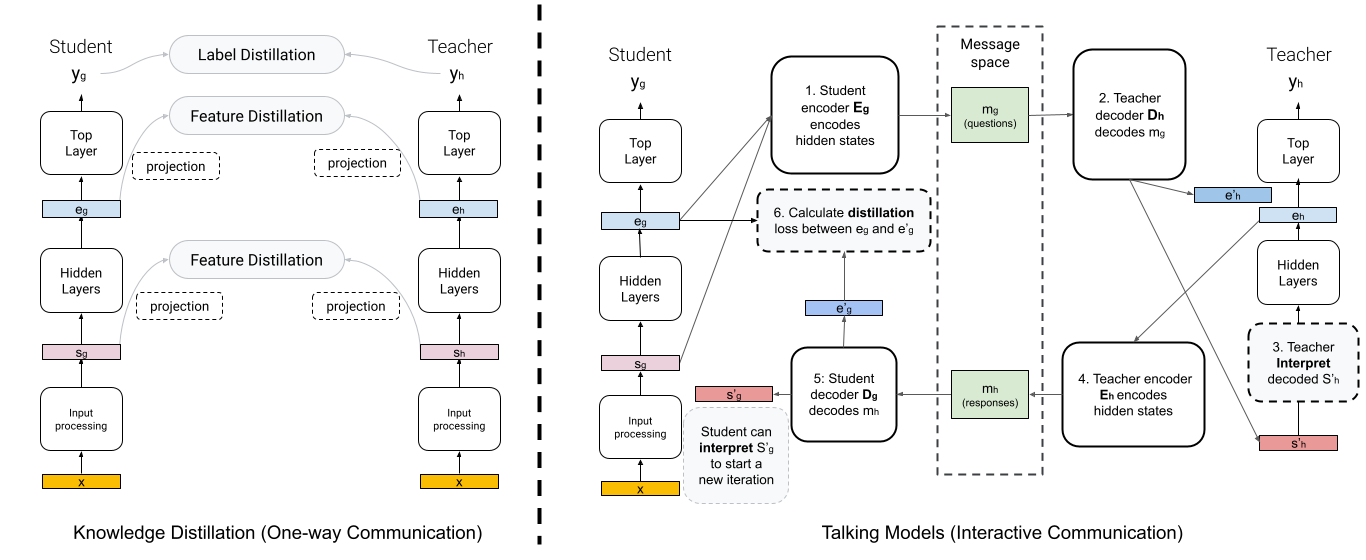}
    \caption{Talking-model Distillation (TD, right) compared to Knowledge Distillation (left, which can be label distillation \citep{hinton2015distilling}, feature distillation \citep{heo2019comprehensive, romero2014fitnets} or hybrid of both \citep{zhu2021student}). In TD, student $M_g$ encodes its hidden states given input $x$ to message $m_g$ with $E_g$, teacher $M_h$ then decodes it with $D_h$, and interprets the decoded hidden states with its hidden layers. $M_h$ encodes the interpreted states to message $m_h$ with $E_h$, Then $M_g$ decodes it with $D_g$ and use it for distillation. $M_g$ can also interpret decoded hidden states and generates ``follow-up'' messages to communicate for multiple iterations.}
    \label{fig:framework}
\end{figure}

We hypothesize that the challenges arise when the pre-trained teacher do not have a good understanding of the student's capacity and ability in learning downstream tasks. We view KD as a one-way communication process between the teacher and student: the teacher generates information (e.g., logits, hidden states) from training data and passes it to the student. And one-way communication can be sub-optimal for knowledge transfer since teacher has no information about the student. In this paper, we extend the KD framework by introducing a two-way communication process, where teacher can provide specific guidance by interpreting student's request. This is inspired by how human communicate and learn from each other: persons from different background and expertise can exchange knowledge effectively by interactive conversations; and students ask question and learn from teacher's response and continue asking follow-up questions. This can be formulated as the Osgood and Schramm Model of communication \citep{wrench2020interpersonal} (shown in Figure \ref{fig:communication}), which is an interpersonal interaction model indicating that messages can go in two directions with a heightened focus on cyclical feedback.

We adapt the Osgood and Schramm Model to design the interactive communication process for KD, where students can ask questions to teachers and learn from the answers. Our proposed process, named Talking-model Distillation (TD), is shown in Figure \ref{fig:framework} (on the right, compared to KD on the left). To be more specific, for each model (i.e., the teacher or student), we introduce two modules used for communication: an encoder that encodes model's hidden states to a message, which will be passed to the other model; and a decoder that decodes the message to its hidden states. 
The communication process starts with student model encoding its hidden states given an input to a message. Teacher then decodes the message into hidden states in its own hidden space. We introduce an interpreting step where teacher runs the decoded hidden states (for bottom layers) using its own learned weights to generate new hidden states. Teacher then encodes the interpreted hidden states into message and returns it to student. Student uses the returned message for knowledge distillation by aligning the decoded hidden states in the returned message with its own hidden states. The training of the encoder and decoder for both teacher and student can happen at the same time as training the student, while the learned weights of the teacher will not be updated. Given the encoder and decoder for each model can be as simple as projection layers using a small number of parameters just to align the corresponding hidden space to a shared message space, the learning of communication parameters does not add much overhead in training the student model. 

By introducing this interactive communication process, the student can request information from teacher based on its own hidden states generated from the downstream tasks. The information teacher provides through the message space is better aligned with downstream tasks by both models' encoders and decoders. Moreover, student can also interpret teacher's returned message and generate a new message back to teacher, as a follow-up. This enables the cyclical communication behavior where multiple iterations of communication can happen, so that student can get sufficient tailored guidance from teacher, even when downstream tasks have extremely sparse training data. 

To show that our proposed method can be used in different domains and applications, we conduct experiments on multiple benchmark datasets including computer vision tasks and recommendation tasks. Compared with state-of-the-art distillation methods (e.g., a hybrid of feature and label distillation \citep{zhu2021student}), our interactive communication process significantly improves knowledge transfer between pre-trained teacher and students of downstream tasks.



\section{Related work}
In this paper, instead of improving general KD, we focus on a specific use-case: use pre-trained large models to teach smaller students for downstream tasks, without extensive fine-tuning of the teacher. Therefore, many most recent KD algorithms cannot be directly applied, as they assume similar tasks between teachers and students \citep{yang2022vitkd, beyer2022knowledge}. In this section, we discuss existing research for efficient model tuning, including model compression and KD, and their inspirations to improve efficiency for downstream models in the pre-training fine-tuning paradigm. 

\subsection{Model compression for efficient serving}
There exist many widely studied research directions to reduce the serving cost of deep neural networks, including but not limited to model pruning \citep{hassibi1993optimal}, quantization \citep{guo2018survey}, and knowledge distillation \citep{hinton2015distilling}. Recent model pruning methods \citep{frantar2022optimal, el2022data, kwon2022fast} have considered large pre-trained models and potential sparse downstream tasks. In these methods, sub-modules of the pre-trained models can be pruned based on their importance towards a pre-defined metric \citep{frantar2022optimal, el2022data}, or masks can be learned to mask out learned weights given downstream tasks \citep{kwon2022fast}. While pruning methods need to be designed based on different model architectures, quantization \citep{krishnamoorthi2018quantizing, guo2018survey} can be applied to any learned weights by replacing them with low-precision data-types. 

Both pruning and quatization directly modify the learned weights of pre-trained models, however, a model with a different structure can be desirable for downstream tasks. Both techniques cannot be applied to newly initialized weights. Knowledge Distillation provides such flexibility and can be complimentary to be used together with pruning and quatization methods \citep{shin2019empirical}. 

\subsection{Knowledge distillation}
Knowledge Distillation uses knowledge from a larger, more powerful ``teacher'' model to improve the performance of a smaller, more efficient ``student'' model \citep{gou2021knowledge}. In recent years, KD has been used in many applications, such as NLP \citep{sun2019patient, sanh2019distilbert}, CV \citep{park2019relational, tung2019similarity}, and Recommendation Systems \citep{tang2018ranking}. 

However, recent research discoveries in understanding knowledge distillation \citep{wang2022efficient, huang2022knowledge, zhu2022teach} suggest a larger teacher does not necessarily guarantee a better student. On the contrary, a huge capacity gap between teachers and students can lead to little or no improvement for the students. This is because some specific representations can only be learned with a large capacity model \citep{zhu2022teach}, or an information bottleneck can be created due to the capacity gap \citep{wang2022efficient}. This is especially challenging when distilling knowledge from large pre-trained foundation models. 

To deal with the capacity gap, existing works introduce intermediate steps/models such as teaching assistant models \citep{mirzadeh2020improved}. Additionally, they try to leverage more information from teachers besides their predictions (or logits), such as feature representations \citep{heo2019comprehensive, yang2019snapshot, zhu2021student, romero2014fitnets}, relationships among labels and representations \citep{hao2022learning, huang2022knowledge}, or the weights of teacher models \citep{fu2021interactive}. 

In this paper, we discuss a specific case of distilling pre-trained foundation models for downstream tasks, where downstream tasks can be different from pre-trained tasks. Fine-tuning a pre-trained foundation model for each downstream task to obtain a specific teacher model can be costly and cannot scale up to many different downstream tasks. 
Existing research in cross-task distillation \citep{yang2022cross, clark2019bam, zhong2022meta} shows that a teacher model trained from multiple tasks can be potentially helpful to a single task student \citep{clark2019bam}. However, for downstream tasks which are not present in teacher models' pre-trained tasks, auxiliary tasks will be needed for both the teacher and student \citep{yang2022cross}. Therefore, we focus on directly distilling information from pre-trained models to students, using their feature representations in hidden spaces similar to the feature distillation techniques \citep{zhu2021student}. 

\subsection{Improve foundation model efficiency for downstream tasks}
The successes of large pre-trained foundation models inspire researchers to explore techniques to improve the efficiency of using them in real-world applications. Many existing research has been focusing on how to improve the efficiency in training or fine-tuning the foundation models. This includes adaptor based method \citep{houlsby2019parameter}, which freezes most of the learned weights and only trains a small adaptor layer for each downstream task. More recently, parameter-efficient fine-tuning techniques extend adaptors to low-rank adaptors which can further improve the training efficiency while reducing serving latency \citep{hu2021lora, he2022parameter}. However, the serving cost of adaptor based fine-tuned model is still comparable to directly using the foundation models. 

More recently, there were early empirical results \citep{hsieh2023distilling, wu2023distilling} showing that distillation method can be used to distill knowledge from pre-trained foundation models to downstream tasks. In NLP specific tasks, by using pre-trained model to generate rationales and explanations of downstream tasks labels, more information can be passed from teacher to student to improve knowledge transfer \citep{hsieh2023distilling}. For text-image foundation model, empirical results showed that vanilla distillation might not directly improve downstream task unless a fine-tuned teacher is used \citep{wu2023distilling}. Inspired by their discoveries, we target at developing distillation techniques that don't require fine-tuning to downstream tasks and can be applied to different machine learning applications and downstream tasks such as CV and Recommendation Systems.

\section{Proposed method}
In this section, we introduce our proposed method named Talking-model Distillation (TD). Our key innovation is to generalize the existing KD methods as a one-way communication process and extend it to a two-way interactive communication process. We first describe how to adapt interpersonal human communication process to machine learning models for KD between teacher and student. We then show that existing KD can be categorized as a one-way communication process. Then we will introduce how TD enables the teacher and student to communicate interactively so that the student can learn rich information from the teacher with specific focus on downstream tasks.

\begin{figure}[t!]
    \centering
    \includegraphics[width=0.8\linewidth]{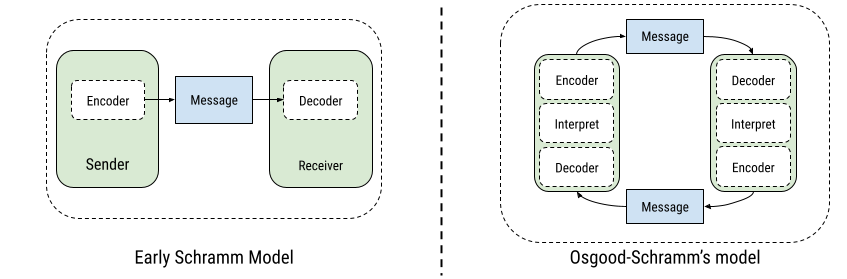}
    \caption{The interpersonal communication models. One-way communication model (left, Early Schramm Model) and interactive communication model (right, Osgood-Schramm's Model)}
    \label{fig:communication}
\end{figure}

\subsection{Communication encoder and decoder}
In Figure \ref{fig:communication}, we show two interpersonal communication models in communication theory \citep{wrench2020interpersonal}, named Early Schramm Model and Osgood-Schramm's Model. Early Schramm Model is a one-way communication model where a sender encodes a message and passes it to a receiver to decode. The Osgood-Schramm's Model is an interactive communication model that two individuals can send, receive and parse messages in a cyclical way, so that the messages being passed can capture feed-backs from each other. 

The key components for both communication processes are: (1) Encoder that encodes information to communicate as messages; (2) Decoder that decodes message to understandable information; (3) A shared message space. To apply these communication models to knowledge distillation, for each of the teacher and student models, we introduce encoder and decoder for communication. 

The first question is what information can encoder encode, while the original KD used logits \citep{hinton2015distilling}, we build encoder on top of the hidden representations of the model, which are more flexible and are not tied to the final prediction tasks \citep{heo2019comprehensive}. Given a model $M_g$ which generates predictions $y_g$ for labels $y$ given input $x$, it typically has multiple hidden layers $H^g_1, ... H^g_{n_g}$ (e.g., Multi-layer perception \citep{gardner1998artificial}, Convolution layers \citep{yamashita2018convolutional}, or Transformer layers \citep{vaswani2017attention}). The encoder can encode any hidden states from these layers to messages. Without loss of generality, we categorize the hidden states into two categories: (1) lower level hidden states $s_g$, which capture the lower-level representations or processed input signals from $H^g_1$ to $H^g_{l_g}$, and (2) higher level hidden states $e_g$, which capture the embeddings that can be used for a few top hidden layers from $H^g_{l_g+1}$ to $H^g_{h_g}$ to generate predictions. These hidden layers for $M_g$ to generate predictions $y_g$ given $x$ can be written as:

$$y_g = H^g_{h_g+1, ..., n_g}(e_g) = H^g_{h_g+1, ..., n_g}(H^g_{l_g+1, ..., h_g}(s_g)) = H^g_{h_g+1, ..., n_g}(H^g_{l_g+1, ..., h_g}(H^g_{1, ... l_g}(x)))$$

Then the encoder $E_g$ for model $M_g$ encodes hidden states from both lower and higher levels into message $m_g$ in the message space $m_g = E_g(\{s_g; e_g\})$. Therefore another model $M_h$ can use its decoder $D_h$ to decode $m_g$ into hidden states for $M_h$: $\{s_h'; e_h'\} = D_h(m_g)$.

The selection of communication encoder and decoder's structure depends on the model architecture as well as the design of the message space. For example, it can be Transformer layers if the model is a transformer and message space contains sequences of embeddings. In this paper, for simplicity, we only explored linear or Dense-Relu-Dense encoders and decoders with message space of $m \in \mathbf{R}^{m_d}$, where $m_d$ is the dimensionality of the message space.

\subsection{Existing KD techniques as one-way communication}

Next we show how we can formulate existing KD techniques as a one-way communication process with the encoder and decoder introduced in the previous subsection. KD techniques here can refer to logit distillation \citep{hinton2015distilling}, feature distillation \citep{romero2014fitnets, heo2019comprehensive} or a hybrid of both \citep{zhu2021student}. We denote student model as $M_g$ and teacher model as $M_h$

For logits distillation, we can simply view the logit space as the message space given both teacher and student having the same prediction tasks. And the distillation loss is $L_{logit} = d(y_g, y_h)$, where d is a distance metric such as L2 loss or KL divergence between the teacher and student logits. 

For feature distillation method (such as \citep{heo2019comprehensive}) which projects hidden states of different dimensionalities into a shared space, i.e., the shared message space, the distillation loss can be calculated by distance measurement between messages in the message space. 

$$L_{feature} = d(m_g, m_h) = d(E_g(\{s_g; e_g\}), E_h(\{s_h; e_h\})) =$$
$$d(E_g(\{H^g_{1, ... l_g}(x); H^g_{l_g+1, ..., h_g}(H^g_{1, ... l_g}(x))\}), E_h(\{H^h_{1, ... l_h}(x); H^h_{l_h+1, ..., h_h}(H^h_{1, ... l_h}(x))\}))$$

Here $d$ is the distance between two messages, which can be L2, KL-divergence or any of the relational metrics that have shown promising results for feature distillation such as Pearson correlation \citep{huang2022knowledge} or manifold loss \citep{hao2022learning}. Encoders for $E_g$ and $E_h$ are learned parameters for feature distillation. 

Another slightly different feature distillation method, Fitnet \citep{romero2014fitnets}, which projects teacher model hidden states to student model's hidden states, can be viewed as student model $M_g$ directly decoding teacher model $M_h$'s hidden states (as message, where $E_h$ is an identity transformation), hence the KD loss can be written as:

$$L_{fitnet} = d(\{s_g; e_g\}, D_g(m_h)) = d(\{s_g; e_g\}, D_g(\{s_h; e_h\})) = $$
$$d(\{H^g_{1, ... l_g}(x); H^g_{l_g+1, ..., h_g}(H^g_{1, ... l_g}(x))\}, D_g(\{H^h_{1, ... l_h}(x); H^h_{l_h+1, ..., h_h}(H^h_{1, ... l_h}(x))\}))$$

where $D_g$'s parameters are learned. We can see that $L_{fitnet}$ distills knowledge by minimizing the distance between original student model's hidden states and decoded states from teacher's hidden states. And Feature distillation $L_{feature}$ distills knowledge by minimizing the distance between encoded student's hidden states and teacher's hidden states in a shared message space. 

\subsection{Interactive communication}

Our proposed method, named Talking-model Distillation (TD) uses interactive communication process, which allow both teacher and student model to interpret messages and return new messages. To be more specific, below we describe one iteration of such interactive communication:

\paragraph{Student passes message to teacher}
Student model $M_g$ generates a message $m_g = E_g(\{s_g; e_g\})$ from input $x$ and pass it to teacher model $M_h$. Then teacher model decodes the message into hidden states with its decoder $\{s_h'; e_h'\} = D_h(m_g)$. 

\paragraph{Teacher interprets message and encodes returned message}
We introduce an \textbf{interpreting} step, where teacher model interprets the decoded message by its own learned weights. Teacher model uses the decoded lower-level hidden states $s_h'$ as input to its hidden layers $H^h_{l_h+1} , ..., H^h_{h_h}$ to generate interpreted states $\tilde{e}_h = H^h_{l_h+1, ..., h_h}(s_h')$. Then teacher encodes the interpreted states (along with the decoded lower-level hidden states) as a returned message: $m_h = E_h(\{s_h', \tilde{e}_h\})$. The interpreting step is crucial for the interactive communication process, as it enables messages from teacher encodes information of teacher's knowledge (model parameters) being applied on student's messages.

\paragraph{Student decodes returned message and learns from teacher}
The student model $M_g$ decodes the returned message to its hidden space: $\{s_g'; e_g'\} = D_g(m_h)$. Then student can learn from teacher by minimizing the distance of the decoded states with its original states:

$$L_{interact} = d(\{s_g; e_g\}, \{s_g'; e_g'\}) = d(\{s_g; e_g\}, D_g(E_h(\{s_h'; \tilde{e}_h\})))$$

Here $d$ can be any distance metric used in existing feature distillation techniques. In this paper, we use L2 loss as $d$ for all distillation methods. We can see the key differences between $L_{interact}$ and other distillation losses ($L_{logit}, L_{feature}$, and $L_{fitnet}$), is that we apply teacher's hidden layers and learned weights on interpreted student's messages, instead of input $x$. By doing so, along with the training of both models' encoder and decoder, teacher can provide feedbacks that fit student model's capacity and learned representation space. Note that $e_h'$ is not used to calculate $L_{interact}$ for learning from the teacher. But it is used for training the teacher's decoder in $L_{SC}$ discussed in the next subsection.  

\paragraph{Student interprets and generates follow-up messages}
After receiving the returned message from teacher, student model can also interpret the message $m_h$ and generate an interpreted state: $\tilde{e}_g = H^g_{l_g+1, ..., h_g}(s_g')$. Then student model can encode it with $s_g'$ to $m_h^2 = E_g(\{s_g', \tilde{e}_g\})$. Here $m_h^2$ refers to the second iteration of the message for interactive communication. It will be passed to teacher again to start the next iteration of communication. 

Note that new iteration of communication won't consume new input $x$ and this can continue for as many iterations as possible. By doing so, rich information from teacher model can pass to student based on student's request, even when downstream tasks are extremely sparse. The iterative update scheme here shares similarity to some techniques in Semi-Supervised Learning and Self-Training \citep{xie2020self}, however, our proposed method used in knowledge distillation are more light-weight and only updates the student model with iterations of interactive communications. We explore the number of iterations for interactive communication as a hyper-parameter in our experiment.

To sum up, our proposed communication process is shown in Figure \ref{fig:framework}, compared with other KD methods. By comparing Figure \ref{fig:framework} with Figure \ref{fig:communication}, we can see that the differences between our method and other KD techniques are similar to the differences between the two interpersonal communication models. A detailed algorithm of our proposed method is included in Appendix \ref{sec:sup_algorithm}.

\subsection{Training of the encoder and decoder for TD}
In order to make the interpreting step and interactive communication effective, it is important for the encoder and decoder to learn a reasonable aligned projection between the message space and each of the model's hidden space. 
This means that given the same input to both teacher and student, encoders of both models need to generate similar messages, and decoded states need to be similar to their original states. To achieve this, besides the $L_{interact}$ used to train student model as well as both student and teacher's encoders and decoders, we introduce following two consistency losses.


\paragraph{Message space consistency} Given the same input $x$, student model $M_g$'s encoder $E_g$ and teacher model $M_h$'s encoder $E_h$ will generate similar messages, described below as the message consistency loss between the two messages. This is similar to $L_{feature}$ for feature distillation. Message consistency loss will be used to train the encoder of both models. 

$$L_{MC} = d(m_g, m_h) = d(E_g(s_g, e_g), E_h(s_h, e_h))$$

\paragraph{State space consistency}
Given the same input $x$, the hidden states decoded by two model's message should be consistent with its own hidden states. We introduce the state consistency loss between decoded states and original states below. This is similar to $L_{fitnet}$ in FitNet and is used to train both encoder and decoder of each model. 

$$L_{SC} = d(\{s_g; e_g\}, D_g(m_h)) + d(\{s_h; e_h\}, D_h(m_g))$$

The communication encoder and decoder will be co-trained with the training of student model, using the combined loss below:

$$L(x, y, M_g, M_h) = L(y, y_g) + w_1 L_{interact}+ w_2 L_{MC} + w_3 L_{SC}$$

Where $L(y, y_g)$ is the groundtruth loss, $w_1, w_2$ and $w_3$ are hyper-parameters of loss weights. 

Note that even though we introduce 3 losses ($L_{interact}, L_{MC}$ and $L_{SC}$), the parameters we added for encoder and decoder are comparable to other feature distillation methods. During training, we freeze teacher model and only update student model and the encoder and decoder of both models. Therefore our method doesn't add notable more weights to learn. However, our distillation process takes more time if the number of iterations of interactive communication is larger than one, due to both teacher and student models need to interpret each other's input multiple times. 

\begin{table*}[t!]
\footnotesize
\centering
\begin{tabular}{l|cccccc} \toprule
 \textbf{Methods}    & \textbf{ML(Dense)} & \textbf{ML(Sparse)} & \textbf{CIFAR10}  & \textbf{CIFAR100}  & \textbf{ImageNet}  & \textbf{Avg.}\\ \midrule
 Train from Scratch & \multicolumn{6}{c}{(---baseline to calculate relative improvement---)}\\ \midrule 
 \textbf{LD} & +0.16\%& +1.49\% & +0.03\% & +1.86\%& -0.21\%  & +0.83\%\\ 
 \textbf{FD} & +0.29\%& +2.68\% & -0.12\% & -0.08\%& -0.14\%  & +0.66\% \\
 \textbf{FitNet} & +0.81\%& +2.19\% & -0.09\% & +0.53\%& +0.29\% & +0.93\% \\ 
 \textbf{Hybrid} & +0.93\%& +2.91\% & +0.23\% & +1.94\% & +0.02\% & +1.50\%\\ \midrule
 Our Method (\textbf{TD}) & \textbf{+1.34\%}& \textbf{+3.39\%} & \textbf{+0.45\%} & \textbf{+2.41\%}& \textbf{+2.56\%} & \textbf{+2.54\%}\\  \bottomrule
\end{tabular}
\caption{Relative improvement of different distillation methods compared to a student model without distillation. Detailed results with standard error are shown in Appendix~\ref{sec:sup_res}.}
\label{tab:e2e}
\end{table*}

\begin{table*}[t!]
\footnotesize
\centering
\begin{tabular}{l|cccccc} \toprule
 \textbf{Methods}    & \textbf{ML(Dense)} & \textbf{ML(Sparse)} & \textbf{CIFAR10}  & \textbf{CIFAR100}  & \textbf{ImageNet}  & \textbf{Avg.}\\ \midrule
 Train from Scratch & \multicolumn{6}{c}{(---baseline to calculate relative improvement---)}\\ \midrule 
 No Interaction & +0.81\%& +2.02\% & +0.43\% & +2.26\% & +2.44\% & +1.99\%\\ 
 1 iteration & +1.31\%& +3.08\% & +0.42\% & +2.31\% & +2.49\% & +2.40\%\\ 
 >1 iterations & \textbf{+1.34\%} & \textbf{+3.30\%} & \textbf{+0.45\%} & \textbf{+2.41\%}& \textbf{+2.56\%} & \textbf{+2.52\%}\\  \bottomrule
\end{tabular}
\caption{Relative improvement of using interactive communication. Detailed results with standard error are shown in Appendix~\ref{sec:sup_res}.}
\label{tab:ablation}
\end{table*}

\section{Experiment}

In this section, we evaluate whether our proposed method can improve knowledge distillation by introducing the interactive communication process, for the case of distilling pre-trained teacher to smaller models for downstream tasks. The teacher model is trained with (multiple) pre-training task(s) or multiple datasets. And the student model will be trained and evaluated on a much smaller dataset with a specific downstream task. 

\subsection{Experiment setup}

\paragraph{Datasets}
We choose multiple widely-used real-world benchmark datasets, including MovieLens \citep{harper2015movielens}, CIFAR10, CIFAR100 \citep{Krizhevsky09learningmultiple}, and ImageNet \citep{ILSVRC15}. These datasets cover applications of recommendation and image classification. For MovieLens (ML), we split the data by timestamps, so that 90\% of the past events will be used to train models evaluated by the 10\% of the future events, which is close to the real-world setup. The pretrained task is to predict movie ratings given a user and a movie for all genres of movies. And downstream tasks are movie rating prediction for a specific genre. Teacher is an MLP model and student model has only 1/4 of the neurons for each layer in teacher model. For image classification datasets, we adopt the same setup as Vision Transformer (ViT) \citep{dosovitskiy2020image}, where we pretrain a large ViT teacher model on ImageNet21K and evaluate student models using CIFAR10, CIFAR100 and ImageNet. Teacher is the pre-trained 12-layer ViT-B/32 model, and student model is the same architecture but only has 4 transformer layers. Dataset details are in Appendix~\ref{sec:sup_dataset}.

\paragraph{Baseline methods}
We compare with four baseline methods: Label Distillation (\textbf{LD} \citep{hinton2015distilling}), Feature Distillation (\textbf{FD} \citep{heo2019comprehensive}), \textbf{FitNet} \citep{romero2014fitnets} and a Hybrid version of Label and Feature Distillation (\textbf{Hybrid} \citep{zhu2021student}).  Note that most recent KD approaches (such as \cite{beyer2022knowledge}, \cite{yang2022cross}) focus on one single application such as image classification or recommendation, and assume teacher and student share similar tasks. We cannot directly apply them in our setup, therefore we compare with the general KD algorithms that can be extended to our use case.
For a fair comparison, all methods use the same teacher and student structure. Each of the methods' KD loss weights are tuned. Details about baseline methods and Hyper-parameter tuning are shown in Appendix~\ref{sec:sup_hp}.

\subsection{Overall improvement}
The overall improvement of our method compared to baseline methods is shown in Table \ref{tab:e2e}. We can see that our method outperforms all baseline methods on all 5 tasks. For MovieLens, we report results on two types of downstream tasks: ML(Dense) is from one genre that has dense data and ML(Sparse) contains 4 genres that are much sparser. We include results on all other genres in Appendix~ \ref{sec:sup_res}.



\subsection{Ablation study: understand the interactive communication}
\label{sec: ablation}
To evaluate the performance of the interactive communication process, we conduct ablation study by adjusting number of iterations for interactive communication. Results are shown in Table \ref{tab:ablation}. For the ``No Interaction'' row, we disable interactive communication (no $L_{interact}$) but keep the encoder and decoder with consistency losses ($L_{MC}$ and $L_{SC}$). We use 3 as the maximum number of iterations for MovieLens and 2 for ViT. The best results are achieved with the maximum number of iterations. We didn't explore larger number of iterations since our training time will be $k$ times slower where $k$ is the number of iterations. But in real-world scenarios, the distillation cost can be relatively small when downstream tasks are sparse. We also conduct ablation studies on the consistency losses $L_{MC}$ and $L_{SC}$ to show the importance of aligning the message spaces and decoded states between the two models. The results are shown in Appendix~\ref{sec:sup_res}. 

\begin{figure}[t!]
    \centering
    \includegraphics[width=0.95\linewidth]{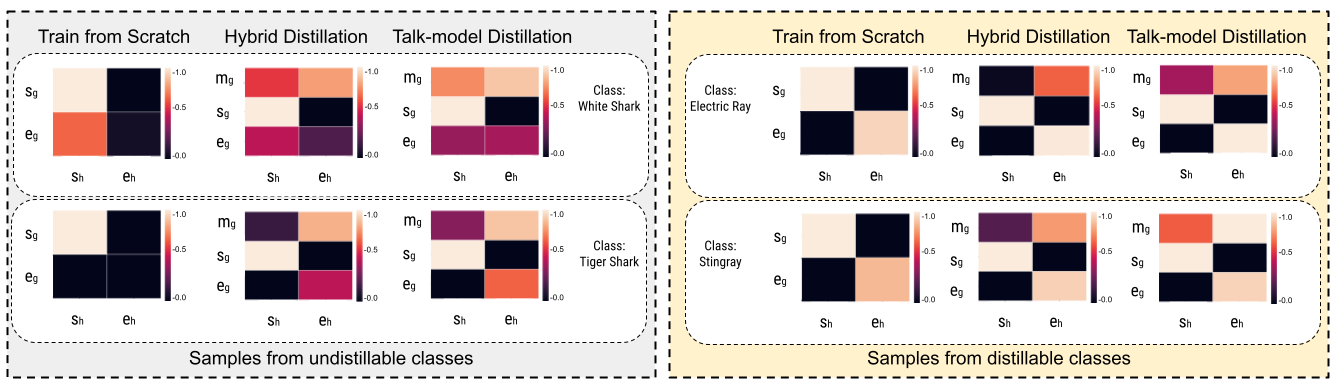}
    \caption{Representation Similarity between teacher (column) and student (row).}
    \label{fig:casestudy}
\end{figure}

\subsection{Cast study: bridge the capacity gap on undistillable downstream classes}
To show that TD can improve student's performance when there is a huge capacity gap between teacher and student, we conduct case study on ImageNet by measuring the representation similarity between student model and teacher model on downstream classes. Similar to the analysis measuring ``undistillable'' classes (classes cannot be distilled) due to capacity gap \citep{zhu2022teach}, we use the Center Kernel Alignment (CKA) \citep{kornblith2019similarity} to measure similarities between representations from teacher ($s_h$ and $e_h$) and student ($s_g$, $e_g$ and the message space $m_g$). For each class, we use 20 examples for the analysis. The results are shown in Figure \ref{fig:casestudy}. The higher the value (i.e., the lighter the color) on the diagonal term, the better aligned the two representation spaces are. We can see that the learned message space is more aligned with teacher compared to distillation baseline even on ``undistillable'' classes (classes with darker color on higher layer representation similarities between teacher and student, i.e., between $e_g$ and $e_h$). And the student models learned with TD are more aligned with teacher's representation on the top layer hidden states ($e_g$ and $e_h$).


\section{Conclusion}
In this paper, we propose Talking-model Distillation (TD), a new distillation paradigm that enables model to communicate interactively, for transferring the knowledge from large pre-trained foundation models to efficient downstream models. By adapting interpersonal communication models to KD, we first show that existing KD techniques utilize one-way communication from teacher to student. We then design an interactive communication paradigm where teacher and student can exchange messages. In this paradigm, teacher can pass knowledge to student based on student's encoded messages using downstream tasks. And the interactive communication can have multiple iterations, which enables distilling rich information even when downstream data are sparse. We conduct experiments on multiple benchmark datasets and we show that our proposed method outperforms other distillation baselines on improving student performance by distilling pre-trained teacher.

\paragraph{Implication and limitation}
Our proposed method provides a new way of thinking about how models can transfer knowledge and interact with each other. Even though we show early results of interactive communication can further improve model performance for knowledge distillation, the gap between teacher and student is still very large. In the same time, we increase the computation resource used in the communication process, even though it is relatively negligible compared to fine-tuning large foundation models. We also believe that by spending computation resources in communication, we can enable models from different tasks even different modalities to learn from each other more efficiently, without the needs of going over large training data. 



\bibliography{references}
\bibliographystyle{apalike}



\newpage

\section{Appendix}

\subsection{Pseudo-Code and more discussion of TD}\label{sec:sup_algorithm}

We provide the pseudo code of our proposed Talking-model Distillation (TD) method in Algorithm \ref{alg:td}. Note that the state consistency loss $L_{SC}$ and message consistency loss $L_{MC}$ will only be applied on $m_g^0$ and $m_h^0$, which require input $x$ being fed to both teacher and student. These two losses are used to train encoders and decoders. They can be disabled in the later training stage. They can also be optional (ablation results shown in Secion \ref{sec:sup_res}). Without these losses, teacher model doesn't need to access any input data $x$ of downstream tasks. 

\begin{algorithm}[h]
    \caption{Pseudo-Code of the proposed TD}
    \label{alg:td}
\begin{algorithmic}[1]
    \REQUIRE Trained teacher model $M_h$, initialized student model $M_g$ (or a pretrained smaller model). Initialized encoders ($E_g$, $E_h$) and decoders ($D_g$, $D_h$) for both teacher and students. Downstream dataset $D = \{X, Y\}$. $k$ iterations for interactive communication.
    \STATE get $x$ and $y$ from D
    \STATE $L = 0.0$   // Total loss.
    \STATE $y_g, \{s_g; e_g\} = M_g(x)$
    \STATE $m_g^0 = E_g(\{s_g; e_g\})$
    \STATE $y_h, \{s_h; e_h\} = M_h(x)$
    \STATE $m_h^0 = E_h(\{s_h; e_h\})$
    \STATE // Adding state consistency and message consistency loss to train 
    \STATE // communication encoders/decoders.
    \STATE $L = L(y, y_g) + w_2 L_{SC}(m_g^0, D_h, m_h^0, D_g) + w_3 L_{MC}(m_g^0, m_h^0)$
    \FOR{each iteration i in [0, k]}
        \STATE $\{s_h'; e_h'\} = D_h(m_g^i)$  // Teacher decodes message from student.
        \STATE $\tilde{e}_h = H^h_{l_h+1, ..., h_h}(s_h')$  // Teacher interpreting step.
        \STATE $m_h^{i+1} = E_g(\{s_h', \tilde{e}_h\})$  // Teacher encodes returned message.
        \STATE $\{s_g'; e_g'\} = D_g(m_h^{i+1})$  // Student decodes returned message.
        \STATE $L = L + w_1 L_{interact}(\{s_g; e_g\}, \{s_g'; e_g'\})$  // Interactive communication loss for iteration i.
        \STATE $\tilde{e}_g = H^g_{l_g+1, ..., h_g}(s_g')$  // Student interpreting step for next iteration.
        \STATE $\{s_g; e_g\} = \{s_g'; \tilde{e}_g\}$  // set student state for next iteration.
        \STATE $m_g^{i+1} = E_g(\{s_g; e_g\})$  // Student encodes message for next iteration.
    \ENDFOR
    \STATE Use optimizer to update the model via total loss $L$.
\normalsize
\end{algorithmic}
\end{algorithm}

Another design choice we made is to make the communication modules exactly the same between teacher and student. This means teacher encoder and student encoder map their states into the same embedding space: both encoders encode all hidden states (i.e., lower layer and higher layer representations); and decoders decode message to all hidden states. However, for distillation, a specialized design might further improve student performance. For example, message from student to teacher only encodes lower layer representation. (Note that message from teacher to student still needs to encode both lower layer and higher layer representations due to student does both distillation and interpreting). In this paper, we intentionally keep the communication operations the same between teacher and student due to the following reasons.

\begin{itemize}
    \item Simplicity. We want to verify the concept of communication for knowledge distillation (KD) and the effectiveness of introducing interactive communication to KD. Therefore, we adopt the simple design to keep the communication mechanism the same between teacher and student.
    \item Future work of multi-way communication. Communication can also happen between a group of models besides two models. Therefore, it is straight-forward to extend the current communication paradigm (both teacher and student adopt same communication mechanism) to multi-way communication, e.g., multiple teachers, and/or multiple students.
    \item Future work of multi-way transfer learning. The communication algorithm we propose in this paper can be applied not only in knowledge distillation, but also can be used as a generic way for transferring knowledge among models. Therefore, it can be applied to two models where they can learn from each other, e.g., between a large Vision Transformer Model and a large language model. Therefore, communication can be used for transfer learning with the unified design of communication encoder and decoder. 
\end{itemize}

\subsection{Dataset description and experiment setup}
\label{sec:sup_dataset}
\paragraph{MovieLens100k \citep{harper2015movielens}} We use the MovieLens100K dataset included in Tensorflow Dataset \footnote{https://www.tensorflow.org/datasets/catalog/movielens}. It contains 100K movie ratings. We use `user\_id', `movie\_id', and `movie\_title' and `movie\_genres' as features. We split the data by timestamps. The 90\% of the data with earlier timestamps are used for training and 10\% of the data with later timestamps are used for evaluation. The time split for evaluation is more realistic than random split for recommendation tasks, since it can capture problems such as user preference shifting overtime as well as cold-starting for new users. We treat training using data from all genres as pre-training task and training on data with movies from specific genres as downstream tasks. We evaluate downstream tasks for 8 most dense genres (with more than 500 evaluation examples), and report the Root Mean Squared Error(RMSE) for rating prediction. 

\paragraph{CIFAR10 \citep{Krizhevsky09learningmultiple}} We use the CIFAR10 dataset included in Tensorflow Dataset \footnote{https://www.tensorflow.org/datasets/catalog/cifar10}. It contains 60,000 32*32 color images in 10 classes. We use the default training and test split, where there are 50,000 images used for training and 10,000 images used for testing.

\paragraph{CIFAR100 \citep{Krizhevsky09learningmultiple}} We use the CIFAR100 dataset included in Tensorflow Dataset \footnote{https://www.tensorflow.org/datasets/catalog/cifar100}. It contains the same 60,000 23*32 color images as CIFAR10, but in 100 classes. We use the same train and test split as CIFAR10.

\paragraph{ImageNet \citep{ILSVRC15}} We use the ImageNet dataset described in Tensorflow Dataset \footnote{https://www.tensorflow.org/datasets/catalog/imagenet2012}. It contains 1,281,168 images for training, 50,000 images for validation and 100,000 images for test. 

\paragraph{Teacher model}
For MovieLens tasks, the teacher model is a Multi-layer Perceptron \citep{gardner1998artificial}, with input dimension 300 (100 for `user\_id', 100 for `movie\_id', and 50 for `movie\_title' using bag-of-words and 50 for  `movie\_genres' using bag-of-words). It has two relu layers of 512 units and 256 units. The model size is tuned as hyper-parameters with a upper limit of cost measured by number of flops, and the optimal teacher model size is below the upper limit. 

For Image classification tasks, we use Vision Transformer (ViT) \citep{dosovitskiy2020image} pre-trained on ImageNet21k with available code \footnote{https://github.com/google-research/vision\_transformer}, hyper-parameter setups and checkpoints \footnote{https://huggingface.co/google/vit-base-patch32-224-in21k}. The teacher model has 16 transformer layers. 

\paragraph{Student model}
For MovieLens, we set student model size to be 1/4 as teacher model: two relu layers of 128 units and 64 units, where we see significantly quality drop compared to teacher models. For image classification tasks, student model only has 4 transformer layers. We find that using the teacher model's pre-trained weights as initialization for the 4 transformer layers and all other layers results in better and more stable performance compared to random initialization. 

\paragraph{Encoder/Decoder}
For encoder and decoder, we use the dense-relu-dense model architecture, with layer norm and dropout. We didn't do extensive hyper-parameter search to choose the best model size, instead, we manually pick relu layer size and message dimension to match the size between teacher and student model. For MovieLens, encoder and decoder have a relu layer with 256 hidden units, and the message dimensionality is 128. For ViT encoder and decoder have a relu layer with 512 hidden units and the message dimensionality is 512. 

\subsection{Hyper-parameter tuning}
\label{sec:sup_hp}
\paragraph{Model parameter} 
For the teacher model on MovieLens, we tune the model size with a upper limit of cost along with learning rate, dropout rate and number of train steps. We don't tune the size of student model, but tune student model's learning rate, dropout rate and number of train steps. For ViT, we use the reported hyper-parameter setup \citep{dosovitskiy2020image}, with fine-tuning steps set to 20000. 

\paragraph{Baseline methods}
For each of the baseline methods, we tune their KD loss weight combined with the groundtruch loss weight. For Label Distillation \citep{hinton2015distilling} (\textbf{LD}) it is the weight of $L_{logit}$, For Feature Distillation \citep{heo2019comprehensive} (\textbf{FD}), it is the weight of $L_{feature}$. For \textbf{FitNet} \citep{romero2014fitnets}, it is the weight of $L_{fitnet}$. And for Hybrid Distillation \citep{zhu2021student} (\textbf{Hybrid}), we tune the weights of overall $L_{logit}$ and $L_{feature}$ and report the best results. 

\paragraph{Our method}
We tune the three weights $w_1$, $w_2$ and $w_3$ for our method, which corresponds to the weight of $L_{interact}$, $L_{SC}$ and $L_{MC}$. We also tune the number of iterations for interactive communication. For MovieLens, it is 0, 1, 2 or 3. And for ViT it is 0, 1 or 2. We report the results with different iteration numbers in our ablation study in Section \ref{sec: ablation}. 

\subsection{Computation resources}
The training of MovieLens can be done on a CPU machine with less than 12 hours for all methods. And the finetuning of ViT models runs on a 4-chip TPU, where all methods finish fine-tuning in 12 hours.    

\begin{table*}[t]
\footnotesize
\centering
\begin{tabular}{l|cccc} \toprule
 \textbf{Methods}    & \textbf{Genre 1} & \textbf{Genre 2} & \textbf{Genre 3}  & \textbf{Genre 4} \\ \midrule
 Train from Scratch & $1.0102 \pm 0.0003$ & $1.0884 \pm 0.0004$ & $1.0555 \pm 0.0000$ & $1.0502 \pm 0.0000$ \\  
 Teacher & $1.0120 \pm 0.0000$ & $1.0854 \pm 0.0000$ & $1.0563 \pm 0.0000$ & $1.0689 \pm 0.0000$ \\\midrule 
 \textbf{LD} & $1.0083 \pm 0.0000$ & $1.1029 \pm 0.0000$ & $1.0601 \pm 0.0000$ & $1.0826 \pm 0.0000$\\ 
 \textbf{FD} & $1.0075 \pm 0.0009$ & $1.0997 \pm 0.0030$ & $1.0537 \pm 0.0007$ & $1.0791 \pm 0.0005$ \\
 \textbf{FitNet} & $1.0018 \pm 0.0000$ & $1.1018 \pm 0.0000$ & $1.0534 \pm 0.0000$ & $1.0849 \pm 0.0000$\\ 
 \textbf{Hybrid} & $1.0015 \pm 0.0008$ & $1.0955 \pm 0.0013$ & $1.0517 \pm 0.0012$ & $\mathbf{1.0737 \pm 0.0014}$ \\ 
 Our Method (\textbf{TD}) & $\mathbf{0.9965 \pm 0.0001}$ & $\mathbf{1.0908 \pm 0.0002}$ & $\mathbf{1.0475 \pm 0.0001}$ & $1.0761 \pm 0.0002$ \\\bottomrule
\end{tabular}
\begin{tabular}{l|cccc} \toprule
 \textbf{Methods}    & \textbf{Genre 5} & \textbf{Genre 6} & \textbf{Genre 7}  & \textbf{Genre 8} \\ \midrule
 Train from Scratch & $1.1609 \pm 0.0000$ & $1.1160 \pm 0.0000$ & $1.0038 \pm 0.0000$ & $1.0937 \pm 0.0000$\\  
 Teacher & $1.1501 \pm 0.0000$ & $1.1090 \pm 0.0000$ & $1.0193 \pm 0.0000$ & $1.0660 \pm 0.0000$\\\midrule 
 \textbf{LD} & $1.1964 \pm 0.0000$ & $1.1415 \pm 0.0000$ & $1.0088 \pm 0.0000$ & $1.0602 \pm 0.0000$\\ 
 \textbf{FD} & $1.1929 \pm 0.0013$ & $1.1269 \pm 0.0029$ & $1.0024 \pm 0.0000$ & $1.0527 \pm 0.0036$ \\
 \textbf{FitNet} & $1.1873 \pm 0.0000$ & $1.1260 \pm 0.0000$ & $1.0032 \pm 0.0000$ & $1.0605 \pm 0.0000$\\ 
 \textbf{Hybrid} & $1.1843 \pm 0.0008$ & $1.1215 \pm 0.0028$ & $1.0076 \pm 0.0009$ & $1.0499 \pm 0.0014$\\ 
 Our Method (\textbf{TD}) & $\mathbf{1.1656 \pm 0.0010}$ & $\mathbf{1.1050 \pm 0.0000}$ & $\mathbf{1.0022 \pm 0.0000}$ & $\mathbf{1.0485 \pm 0.0013}$\\\bottomrule
\end{tabular}
\caption{RMSE of rating prediction on MovieLens Genre 1 to Genre 8 (Dense to Sparse), compared to baseline methods. \textbf{bold} numbers for the best improvement given a certain genre.}
\label{tab:ml_genre}
\end{table*}

\begin{table*}[t]
\footnotesize
\centering
\begin{tabular}{l|ccc} \toprule
 \textbf{Methods}    & \textbf{CIFAR10} & \textbf{CIFAR100} & \textbf{ImageNet}  \\ \midrule
 No Distillation & $0.93678$ & $0.74764$ & $0.48683$ \\   \midrule
 \textbf{LD} & $0.93709$ & $0.76162$ & $0.48576$ \\ 
 \textbf{FD} & $0.93565$ & $0.74702$ & $0.48615$ \\
 \textbf{FitNet} & $0.93586$ & $0.75164$ & $0.48828$ \\ 
 \textbf{Hybrid} & $0.93894$ & $0.76213$ & $0.48691$ \\ 
 Our Method (\textbf{TD})  & $\mathbf{0.94100}$ & $\mathbf{0.76562}$ & $\mathbf{0.49930}$ \\\bottomrule
\end{tabular}
\caption{Accuracy of image classification tasks, compared to baseline methods. \textbf{bold} numbers for the best results on a dataset.}
\label{tab:vit_per_task}
\end{table*}

\subsection{Additional experimental results}
\label{sec:sup_res}
In this subsection, we include detailed experimental results. For experiments on MovieLens, both teacher and student models are random initialized. For each result, we run the same setup 5 times and report the mean RMSE with standard error. For experiments using image classification tasks, teacher is pre-trained ViT and students are initialized using the learned weights from pre-trained teacher (only the first four transformer layers), therefore the results have low variance and we only run each setup once due to the limit of computation resources.

\paragraph{MovieLens100k per genre results}

Results (RMSE, lower is better) on MoiveLens are shown in Table \ref{tab:ml_genre}. From where we can see that our method outperforms baseline methods on 7 of the 8 genres. We also include the results of the teacher model, which is trained on all genres. The teacher model is not fine-tuned to each downstream genre, and different genres can have very different data distributions. Therefore, in some genres a model trained from scratch is better than the teacher model. And distillation from teacher to student could even hurt the student's performance for some genres. This real challenge in recommendation tasks and many other downstream applications inspires us to design the interactive communication process so that knowledge aligned with downstream tasks can be transferred effectively. We can see that our method, though does not improve the student model on some specific genres, can out-perform both teacher and student on most genres.

\paragraph{Vision Transformer results}

Results (classification accuracy, higher is better) for image classification tasks are shown in Table \ref{tab:vit_per_task}. We can see that our method outperforms baseline methods on all downstream tasks. Our improvement is most significant on ImageNet, which is a much more difficult task compared to CIFAR10 and CIFAR100. Note that the pre-trained teacher cannot be directly applied to downstream tasks, due to classification label mismatch, so we don't report teacher's results. However, the fine-tuned results can be found in the ViT paper \citep{dosovitskiy2020image} (0.98, 0.92, 0.81 for CIFAR10, CIFAR100 and ImageNet). We can see that there is still a huge gap between teacher and student. 

We want to point out that in this paper we don't discuss the upper limit of the student model nor try to close the gap between teacher and student. In our case, we expect the student with 4 transformer layers to perform much worse than the teacher with 12 transformer layers. We want to verify that by using the proposed interactive communication process, we can transfer more useful knowledge from a powerful pre-trained foundation model to much smaller models for downstream applications, compared to existing KD baseline methods.


\begin{table*}[t]
\footnotesize
\centering
\begin{tabular}{l|cccc} \toprule
 \textbf{Methods}    & \textbf{Genre 1} & \textbf{Genre 2} & \textbf{Genre 3}  & \textbf{Genre 4} \\ \midrule
 No Interactions & $1.0018 \pm 0.0002$ & $1.1016 \pm 0.0002$ & $\mathbf{1.0475 \pm 0.0001}$ & $1.0767 \pm 0.0000$\\
 1 iteration & $0.9971 \pm 0.0003$ & $\mathbf{1.0908 \pm 0.0002}$ & $1.0502 \pm 0.0007$ & $1.0783 \pm 0.0002$\\
 2 to 3 iterations & $\mathbf{0.9965 \pm 0.0001}$ & $1.0921 \pm 0.0005$ & $1.0508 \pm 0.0002$ & $\mathbf{1.0761 \pm 0.0002}$\\ \bottomrule
\end{tabular}
\begin{tabular}{l|cccc} \toprule
 \textbf{Methods}    & \textbf{Genre 5} & \textbf{Genre 6} & \textbf{Genre 7}  & \textbf{Genre 8} \\ \midrule
 No Interactions & $1.1864 \pm 0.0014$ & $1.1249 \pm 0.0000$ & $\mathbf{1.0022 \pm 0.0000}$ & $1.0656 \pm 0.0000$\\ 
 1 iteration & $1.1667 \pm 0.0008$ & $1.1134 \pm 0.0008$ & $1.0033 \pm 0.0002$ & $1.0496 \pm 0.0026$\\
 2 to 3 iterations & $\mathbf{1.1663 \pm 0.0016}$ & $\mathbf{1.1050 \pm 0.0000}$ & $1.0039 \pm 0.0004$ & $\mathbf{1.0495 \pm 0.0004}$\\ \bottomrule
\end{tabular}
\caption{RMSE of rating prediction on MovieLens Genre 1 to Genre 8 (Dense to Sparse), with different number of iterations for interactive communication. \textbf{bold} numbers for the best results given a certain genre.}
\label{tab:ml_iter}
\end{table*}

\begin{table*}[t]
\footnotesize
\centering
\begin{tabular}{l|ccc} \toprule
 \textbf{Methods}    & \textbf{CIFAR10} & \textbf{CIFAR100} & \textbf{ImageNet}  \\ \midrule
 No Interactions & $0.94089$ & $0.76460$ & $0.49873$ \\ 
 1 iteration & $0.94069$ & $0.76511$ & $0.49893$ \\
 2 iterations & $\mathbf{0.94100}$ & $\mathbf{0.76562}$ & $\mathbf{0.49930}$\\\bottomrule
\end{tabular}
\caption{Accuracy of image classification tasks, with different number of iterations for interactive communication. \textbf{bold} numbers for the best results on a dataset.}
\label{tab:vit_iter}
\end{table*}

\begin{table*}[t]
\footnotesize
\centering
\begin{tabular}{l|cccc} \toprule
 \textbf{Methods}    & \textbf{Genre 1} & \textbf{Genre 2} & \textbf{Genre 3}  & \textbf{Genre 4} \\ \midrule
 No $L_{MC}$ & $\mathbf{0.9965 \pm 0.0001}$ & $\mathbf{1.0908 \pm 0.0003}$ & $1.0481 \pm 0.0010$ & $1.0767 \pm 0.0003$\\
 No $L_{SC}$ & $0.9968 \pm 0.0003$ & $1.0916 \pm 0.0004$ & $1.0501 \pm 0.0002$ & $1.0763 \pm 0.0002$\\
 Our Method(\textbf{TD}) & $\mathbf{0.9965 \pm 0.0001}$ & $\mathbf{1.0908 \pm 0.0002}$ & $\mathbf{1.0475 \pm 0.0001}$ & $\mathbf{1.0761 \pm 0.0002}$\\ \bottomrule
\end{tabular}
\begin{tabular}{l|cccc} \toprule
 \textbf{Methods}    & \textbf{Genre 5} & \textbf{Genre 6} & \textbf{Genre 7}  & \textbf{Genre 8} \\ \midrule
 No $L_{MC}$  & $1.1683 \pm 0.0007$ & $1.1104 \pm 0.0007$ & $1.0035 \pm 0.0000$ & $\mathbf{1.0485 \pm 0.0013}$\\ 
 No $L_{SC}$ & $1.1658 \pm 0.0012$ & $1.1058 \pm 0.0002$ & $1.0034 \pm 0.0001$ & $1.0494 \pm 0.0004$\\
 Our Method(\textbf{TD}) & $\mathbf{1.1656 \pm 0.0010}$ & $\mathbf{1.1050 \pm 0.0000}$ & $\mathbf{1.0022 \pm 0.0000}$ & $\mathbf{1.0485 \pm 0.0013}$\\ \bottomrule
\end{tabular}
\caption{RMSE of rating prediction on MovieLens Genre 1 to Genre 8 (Dense to Sparse), ablating different losses. \textbf{bold} numbers for the best results given a certain genre.}
\label{tab:ml_loss}
\end{table*}

\begin{table*}[t]
\footnotesize
\centering
\begin{tabular}{l|cccc} \toprule
 \textbf{Methods}    & \textbf{Genre 1} & \textbf{Genre 2} & \textbf{Genre 3}  & \textbf{Genre 4} \\ \midrule
 Add Noise & $0.9966 \pm 0.0001$ & $1.0910 \pm 0.0002$ & $1.0511 \pm 0.0003$ & $1.0762 \pm 0.0003$\\
 No Noise & $0.9966 \pm 0.0002$ & $1.0912 \pm 0.0004$ & $1.0475 \pm 0.0001$ & $1.0766 \pm 0.0002$\\  \midrule
 Train $E_*,D_*$ separately & $1.0008 \pm 0.0002$ & $1.1066 \pm 0.0001$ & $1.0475 \pm 0.0001$ & $1.0812 \pm 0.0000$\\
 Train together & $0.9965 \pm 0.0001$ & $1.0908 \pm 0.0002$ & $1.0508 \pm 0.0002$ & $1.0761 \pm 0.0002$\\\bottomrule
\end{tabular}
\begin{tabular}{l|cccc} \toprule
 \textbf{Methods}    & \textbf{Genre 5} & \textbf{Genre 6} & \textbf{Genre 7}  & \textbf{Genre 8} \\ \midrule
 Add Noise & $1.1663 \pm 0.0009$ & $1.1052 \pm 0.0001$ & $1.0033 \pm 0.0001$ & $1.0487 \pm 0.0014$\\
 No Noise & $1.1668 \pm 0.0015$ & $1.1051 \pm 0.0002$ & $1.0026 \pm 0.0003$ & $1.0504 \pm 0.0012$\\  \midrule
 Train $E_*,D_*$ separately & $1.1782 \pm 0.0002$ & $1.1165 \pm 0.0002$ & $1.0022 \pm 0.0000$ & $1.0603 \pm 0.0002$\\
 Train together & $1.1656 \pm 0.0010$ & $1.1050 \pm 0.0000$ & $1.0061 \pm 0.0001$ & $1.0485 \pm 0.0013$\\ \bottomrule
\end{tabular}
\caption{RMSE of rating prediction on MovieLens Genre 1 to Genre 8 (Dense to Sparse), by adding noise before teacher's interpreting or separately training encoder/decoder.}
\label{tab:ml_tune}
\end{table*}

\begin{table*}[t]
\footnotesize
\centering
\begin{tabular}{l|ccc} \toprule
 \textbf{Methods}    & \textbf{CIFAR10} & \textbf{CIFAR100} & \textbf{ImageNet}  \\ \midrule
 No $L_{SC}$  & $0.94069$ & $0.76398$ & $0.49930$ \\ 
 Add Noise & $0.94089$ & $0.76562$ & $0.49917$ \\
 No Noise & $0.94100$ & $0.76511$ & $0.49930$ \\
 Our Method(\textbf{TD}) & $0.94100$ & $0.76562$ & $0.49930$\\\bottomrule
\end{tabular}
\caption{Accuracy of image classification tasks, ablating $L_{SC}$ or adding noises before teacher's interpreting.}
\label{tab:vit_ablate}
\end{table*}

\paragraph{Ablation of interactive communication}
We evaluate the effectiveness of interactive communication by changing the number of iterations for calculating the interactive communication loss $L_{interact}$. Results on MovieLens are shown in Table \ref{tab:ml_iter} and results on image classification are shown in Table \ref{tab:vit_iter}. We can see that even without interactive communication, our method can outperform some baseline methods. We think this is because the introduction of both $L_{SC}$ and $L_{MC}$ enables better alignment between the student and teacher's hidden states. It can be viewed as a combination of FitNet and feature distillation. And by introducing interactive communication, we further improve the student model. 

\paragraph{Ablation of consistency losses}
We also evaluate the importance of the consistency losses we introduced to help training the communication encoder and decoder. Results on MovieLens are shown in Table \ref{tab:ml_loss} and results on image classification are shown in Table \ref{tab:vit_ablate}. For MovieLens, we can see that both message consistency loss $L_{MC}$ and state consistency loss $L_{SC}$ are useful for most genres. For ViT, we always add $L_{MC}$ since we observe that without $L_{MC}$ the communication encoder and decoder is hard to train. And we see that $L_{SC}$ improves the model on CIFAR10 and CIFAR100 but not ImageNet. We think applying $L_{interact}$ with multiple iterations can train the communication encoder and decoder reasonable well, therefore the consistency losses may not always be useful on all downstream tasks.

\paragraph{Adding noises during communication}
Inspired by ideas in self-training and semi-supervised learning \citep{xie2020self}, where noise can be added to input or representation to improve the generalization and robustness of knowledge transfer, we also explored the option of adding noise in the interpreting process. Specifically, we add a small Gaussian noise on $s_h'$, which is the decoded lower layer presentation for teacher model to interpret. Results on MovieLens are shown in Table \ref{tab:ml_tune} and results on image classification are shown in Table \ref{tab:vit_ablate}. We can see that adding noise can improve performance on some downstream tasks but not all of them. 

\paragraph{Separate training of encoder/decoder}
We also explored different ways of improving the learning of communication encoder and decoder. One way is to introduce a ramp-up stage where only these encoders and decoders are trained. To do this, we first train student model a few steps (1000 on MovieLens) and then we freeze the student model and only train both teacher and student's encoders and decoders for another few steps (500 or 1000 on MovieLens). We report the results on MoiveLens in Table \ref{tab:ml_tune}, where we can see it does not necessarily improve the student model's performance. One reason is that introducing this ramp-up step will relatively reduce the train steps of end-to-end training. Therefore, it requires more tuning on learning rate, train steps to identify improvement with this training schema. To keep the experiment and algorithm design as simple as possible, in our experiments, we train everything (student model, both teacher and student's encoders and decoders) together in a single stage. 


\end{document}